\documentclass[conference]{IEEEtran}

\usepackage{cite}
\usepackage{amsmath,amssymb,amsfonts}
\usepackage{graphicx}
\usepackage{stfloats}
\usepackage{placeins}
\usepackage{xcolor}
\usepackage{booktabs}
\usepackage{multirow}
\usepackage{makecell}
\usepackage{array}
\usepackage{tabularx}
\usepackage{adjustbox}
\usepackage{url}
\usepackage{xspace}
\usepackage[hidelinks]{hyperref}

\graphicspath{{figures/}}
\newcommand{\modelname}{scKDGM\xspace}

\begin{document}

\title{\modelname: KAN-guided Dynamic Graph Masked Learning for Single-Cell RNA-seq Clustering}

\author{
\IEEEauthorblockN{
Jun Tang\IEEEauthorrefmark{1},
Pengwei Hu\IEEEauthorrefmark{2}\textsuperscript{*},
Sicong Gao\IEEEauthorrefmark{1},
Jie Guo\IEEEauthorrefmark{1},
Lun Hu\IEEEauthorrefmark{2},
Xin Luo\IEEEauthorrefmark{1}
}

\IEEEauthorblockA{
\IEEEauthorrefmark{1}
\textit{College of Computer and Information Science},
\textit{Southwest University},
Chongqing, China \\
Email: swutangjun@email.swu.edu.cn;
gsc020825@email.swu.edu.cn;
myliya46@email.swu.edu.cn;
luoxin@swu.edu.cn
}

\IEEEauthorblockA{
\IEEEauthorrefmark{2}
\textit{Xinjiang Technical Institute of Physics and Chemistry},
\textit{Chinese Academy of Sciences},
Urumqi, China \\
Email: hpw@ms.xjb.ac.cn; hulun@ms.xjb.ac.cn \\
\textsuperscript{*}Corresponding author: Pengwei Hu.
}
}

\maketitle

\begin{abstract}
Single-cell RNA sequencing (scRNA-seq) clustering is essential for identifying cell types, but high dimensionality, sparsity, dropout, and technical noise hinder robust expression representation and cell graph construction. Existing masked autoencoders mainly use expression recovery for feature reconstruction, while graph clustering methods usually depend on fixed KNN graphs and do not feed recovered expression back into graph optimization. We propose \modelname, a KAN-guided dynamic graph masked learning framework for scRNA-seq clustering. \modelname uses graph-aware distribution preserving gene masking (GDP-Mask) to perturb cell identity, a KAN-based TAKGCN encoder to learn masked-view representations, mask-guided expression recovery to construct a dynamic graph, and cross-view contrastive learning to transfer recovery signals into topology updates. A ZINB loss models overdispersion and zero inflation. Experiments on 12 real scRNA-seq datasets show that \modelname outperforms 10 baselines in average NMI and ARI.
\end{abstract}


\section{Introduction}
Single-cell RNA sequencing (scRNA-seq) profiles transcriptomic heterogeneity at single-cell resolution and supports cell type identification, trajectory analysis, and disease microenvironment studies. Early single-cell whole-transcriptome sequencing\cite{Tang2009} and high-throughput platforms such as Drop-seq\cite{Macosko2015} and 10x Genomics\cite{Zheng2017} made large-scale scRNA-seq routine. As data scale and tissue complexity grow, unsupervised cell subpopulation identification remains central to scRNA-seq analysis\cite{Luecken2019,Kiselev2019Challenges}. Related structured-recognition studies in biomedical and visual domains also show the value of attention, tensor decomposition, fuzzy clustering, and multiscale features under noisy observations\cite{SourceList48,SourceList57,SourceList37,SourceList42}.

scRNA-seq data are high-dimensional, sparse, and dropout-prone, so direct clustering on Euclidean distances or low-dimensional projections is sensitive to noise. Seurat\cite{Satija2015}, SC3\cite{Kiselev2017SC3}, and CIDR\cite{Lin2017CIDR} improve stability with nearest-neighbor graphs, consensus clustering, or implicit imputation, but still rely on handcrafted features or shallow assumptions. Deep methods learn richer representations through embedded clustering, ZINB or negative-binomial autoencoding, metric learning, masked estimation, and contrastive learning\cite{Tian2019scDeepCluster,Chen2020scziDesk,Yu2023scDML,Wan2022scNAME,Wang2024CIRCLE,Fang2024scMAE}. Graph methods model cell relations with GNNs, graph autoencoders, adaptive graphs, and graph contrastive objectives\cite{Wang2021scGNN,Gan2022scDSC,Yu2022scTAG,Yu2023scMGCA,Xu2024scCDCG,Wang2025scE2EGAE,Li2025scAGC,Tian2025scGZDC}. Biological-prior graphs in scPriorGraph\cite{cao2024scpriorgraph}, pathway-consensus graphs in scMCGraph\cite{huang2025consensus}, and masked graph autoencoding\cite{Hou2022GraphMAE} further improve representation learning. Broader graph studies similarly emphasize spatiotemporal message passing, high-order filters, global dependency learning, node collaboration, graph pooling, attributed clustering, community search, graph transitions, metapath associations, and graph-regularized factorization\cite{SourceList03,SourceList04,SourceList05,SourceList07,SourceList47,SourceList36,SourceList27,SourceList43,SourceList39,SourceList64}. Yet most graph clustering pipelines still use a fixed KNN graph, which can form hubs, weaken rare or low-degree cells, and turn expression recovery into reconstruction-only supervision.

The expression recovery branch is also related to high-dimensional incomplete (HDI) representation learning. Reviews and adaptive-divergence latent factor models show that objective design matters under sparse observations\cite{SourceList30,SourceList11,SourceList13}. PI/PID control, fuzzy SGD, asynchronous or accelerated parallel SGD, ADMM, Nesterov acceleration, PSO, genetic search, and coevolutionary optimization provide related strategies for stable latent-factor learning\cite{SourceList02,SourceList08,SourceList06,SourceList10,SourceList63,SourceList60,SourceList67,SourceList56,SourceList61,SourceList20,SourceList26,SourceList68}. Nonlinear HDI models based on randomized or nonnegative factors, hash factors, multimetric features, autoencoders, outlier-resilient reconstruction, and prediction sampling further motivate robust feature recovery from noisy high-dimensional matrices\cite{SourceList14,SourceList15,SourceList17,SourceList59,SourceList62,SourceList28,SourceList49,SourceList65}.

The dynamic graph component is motivated by temporal and tensor representation studies. Kalman filtering, temporal QoS modeling, neighborhood regularization, temporal bias, and traffic imputation show how sparse observations can be refined by temporal structure\cite{SourceList01,SourceList12,SourceList16,SourceList31,SourceList38,SourceList53}. Mode-aware Tucker networks, neural Tucker factorization, dynamic graph mixers, attention-based or neural tensor factorization, and battery-life tensor prediction indicate that nonlinear and mode-aware tensor models can capture interactions beyond static features\cite{SourceList19,SourceList25,SourceList40,SourceList52,SourceList22,SourceList34,SourceList41,SourceList50}. Dynamic transaction networks, tensor causal convolution, convolution-bias factorization, fine-grained tensor regularization, electricity-theft detection, and spatiotemporal recovery further support linking recovered signals with evolving relational structure\cite{SourceList29,SourceList32,SourceList33,SourceList55,SourceList18,SourceList51}.

We propose \modelname, a KAN-guided dynamic graph masked learning framework for robust scRNA-seq clustering. \modelname uses a KAN-based Topology Adaptive Graph Convolutional Network (TAKGCN) encoder. KAN replaces fixed activations and scalar linear weights with learnable univariate edge functions\cite{Liu2025KAN}, and KAN-based graph networks extend this idea to graph learning\cite{Bresson2025KAGNN,Li2025KAGNNMI}. TAKGCN also follows TAGCN-style multi-hop aggregation\cite{Du2018TAGCN}; modular graph convolution, graph tensor attention, and network compression studies provide related evidence that graph operators and compact neural transformations can improve structured representation learning\cite{SourceList24,SourceList09,SourceList21}. On this basis, GDP-Mask samples non-neighbor donor cells and shuffles values within each gene column. The model recovers expression from the masked view, builds a differentiable dynamic graph from the recovered matrix, and aligns masked and dynamic views by contrastive learning. Thus expression recovery directly refines graph topology instead of acting only as reconstruction supervision. Our contributions are:
\begin{itemize}
\item First, we propose TAKGCN, which aggregates high-order cell-neighborhood information through topology-adaptive graph convolution and models nonlinear gene-expression dependencies with Fourier KAN-style transformations.
\item Second, we design GDP-Mask to construct graph-aware and distribution preserving cell identity perturbations for masked recovery. 
\item Third, we introduce a mask-recovery-driven dynamic graph learning mechanism that uses recovered expression information to refine cell adjacency. 
\item Experiments on 12 real scRNA-seq datasets show that \modelname achieves the best average NMI and ARI under the benchmark setting.
\end{itemize} 

\section{Methods}
\subsection{Overview}
Given an scRNA-seq matrix \(X\in\mathbb{R}^{n\times g}\), \modelname learns \(Z\in\mathbb{R}^{n\times d}\) and partitions \(n\) cells into \(C\) latent groups. Preprocessing removes zero-count genes and cells, applies library-size normalization, median scaling, log transformation, and Scanpy-based selection of the top 1000 highly variable genes\cite{wolf2018scanpy}. A KNN graph \(A_0\) initializes message passing.

\begin{figure*}[t]
    \centering
    \includegraphics[width=0.95\textwidth]{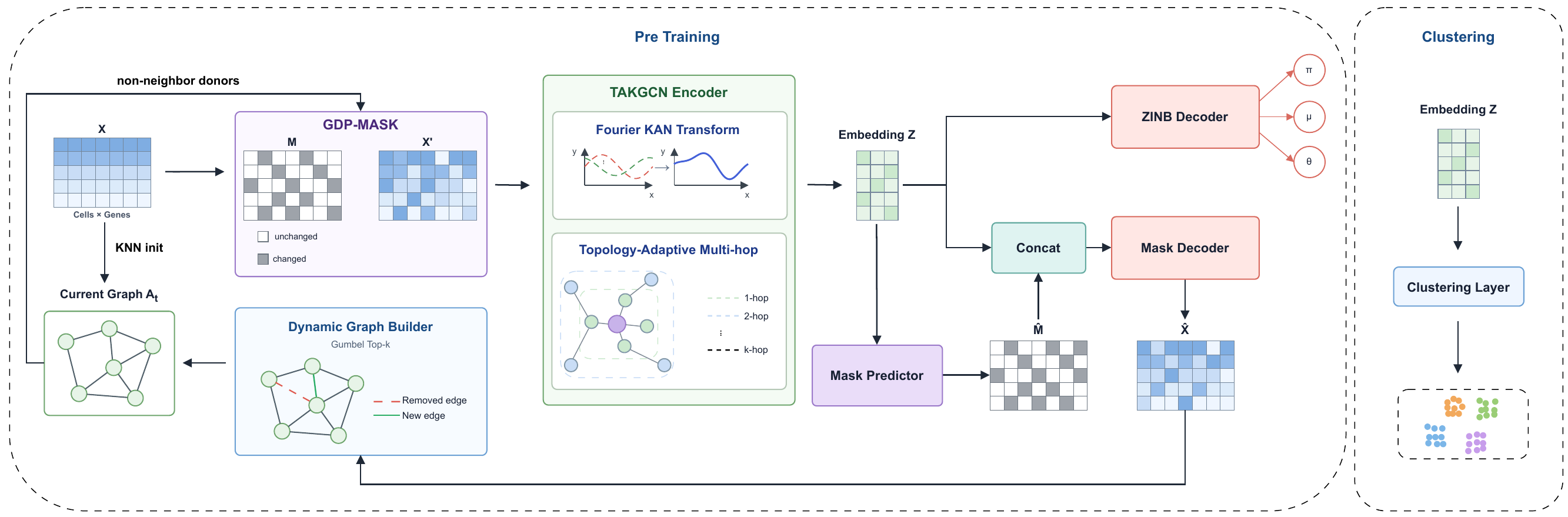}
    \caption{The architecture of \modelname. GDP-Mask produces the graph-aware masked feature matrix and observed-change mask. TAKGCN encodes the masked expression on the current graph, while the recovered feature matrix is used to construct dynamic graph.}
    \label{fig:model}
\end{figure*}

Fig.~\ref{fig:model} shows the two-stage pipeline. During masked pre-training, GDP-Mask generates \(X'\) and an observed-change mask \(M\) from \(X\) and the current graph \(A_t\). TAKGCN encodes \(X'\) on \(A_t\), predicts \(M\), reconstructs \(\hat X\), and uses \(\hat X\) to build a differentiable dynamic graph. Clean expression is re-encoded on this graph and aligned with the masked view. During clustering refinement, the mask branch is removed, DEC-style clustering is optimized on clean expression, and dynamic-graph contrastive learning remains active.

\subsection{GDP-Mask}
GDP-Mask perturbs expression by graph-aware, gene-wise shuffling. For cell \(i\), non-neighbor donors under \(A_t\) form
\begin{equation}
    \mathcal{D}_t(i)=\{r\mid A_t(i,r)=0,\ r\neq i\}.
\end{equation}
If \(\mathcal{D}_t(i)\) is empty, the donor is sampled from non-self cells. With \(\pi_t(i)\sim \mathrm{Uniform}(\mathcal{D}_t(i))\), GDP-Mask samples
\begin{equation}
    S_{ij}\sim \mathrm{Bernoulli}(\rho),
\end{equation}
where \(\rho\) is the mask rate, and replaces only within the same gene column:
\begin{equation}
    X'_{ij} =
    \begin{cases}
        X_{\pi_t(i)j}, & S_{ij}=1\\
        X_{ij}, & S_{ij}=0
    \end{cases}.
\end{equation}
This preserves each gene's marginal distribution while introducing non-neighbor cell identity confusion. The supervised target is the observed-change mask
\begin{equation}
    M_{ij}=\mathbb{I}(X'_{ij}\neq X_{ij}).
\end{equation}
Thus supervision ignores unchanged sampled entries, such as \(0\rightarrow0\), and forces recovery of cell-specific expression from graph context and gene combinations.

\subsection{TAKGCN Encoder}
TAKGCN combines TAG-style multi-hop aggregation with KAN-style nonlinear transformations. Let \(H^{(\ell)}\) be the layer input and \(\tilde A\) the normalized adjacency. One TAKGConv layer is
\begin{equation}
    H^{(\ell+1)}=\sum_{r=0}^{K}\Phi_{r}\left(\tilde A^{r}H^{(\ell)}\right)+b.
\end{equation}
Here \(K\) is the maximum hop number. The hop-specific Fourier KAN transform is
\begin{equation}
    \Phi_r(x)_o=\sum_{i=1}^{d_{\mathrm{in}}}\sum_{q=1}^{G}
    a_{oiq}^{(r)}\left[\cos(qx_i/s)-1\right]+
    b_{oiq}^{(r)}\sin(qx_i/s).
\end{equation}
Here \(G\) is the number of Fourier harmonics and \(s=10\) scales inputs to avoid overly rapid oscillation. The \(\cos(\cdot)-1\) term centers the basis at zero input. This gives TAKGCN both high-order graph receptive fields and flexible nonlinear gene-expression fitting.

\subsection{Recovery-driven Dynamic Graph Learning}
After GDP-Mask, TAKGCN encodes
\begin{equation}
    Z_m=f_{\mathrm{enc}}(X^{\prime},A_t).
\end{equation}
The mask predictor estimates changed entries,
\begin{equation}
    \hat{M}=f_m(Z_m).
\end{equation}
with
\begin{equation}
    \mathcal{L}_{\mathrm{mask}}
    = \mathrm{BCEWithLogits}(\hat M, M).
\end{equation}
The feature decoder uses both \(Z_m\) and \(\sigma(\hat M)\):
\begin{equation}
    \hat{X}=f_x\left(\left[Z_m, \sigma(\hat{M})\right]\right).
\end{equation}
Weighted MSE emphasizes observed-change positions:
\begin{align}
    \mathcal{L}_{\mathrm{rec}}=
    \frac{1}{np}\sum_{i,j} w_{ij}(\hat X_{ij}-X_{ij})^2\\
    w_{ij}=\alpha M_{ij}+(1-\alpha)(1-M_{ij}).
\end{align}
The ZINB decoder provides an auxiliary \(\mathcal{L}_{\mathrm{ZINB}}\) for zero inflation and overdispersion. Crucially, \(\hat X\) also drives graph learning. During pre-training, \(U=\hat X\); during refinement, \(U=Z_c=f_{\mathrm{enc}}(X,A_t)\). After row-normalizing \(U\), the model computes
\begin{equation}
    R_{ij}=\frac{u_i^\top u_j}{\|u_i\|_2\|u_j\|_2},\quad R_{ii}=-\infty .
\end{equation}
With Gumbel noise
\begin{equation}
    G_{ij}=-\log(-\log \epsilon_{ij}),\quad \epsilon_{ij}\sim \mathrm{Uniform}(0,1),
\end{equation}
the soft edge probability is
\begin{equation}
    P_{ij}=\mathrm{softmax}_{j}\left((R_{ij}+G_{ij})/\tau_g\right),
\end{equation}
where \(\tau_g\) is the Gumbel temperature. Top-\(k\) selection yields
\begin{equation}
    A^{\mathrm{hard}}_{ij}
    =
    \mathbb{I}\left(j\in \mathrm{TopK}(P_i,k)\right).
\end{equation}
The straight-through graph keeps a sparse forward pass and soft gradients:
\begin{equation}
    P^{k}=P\odot A^{\mathrm{hard}},
\end{equation}
\begin{equation}
    A^{\mathrm{st}}
    =
    A^{\mathrm{hard}}
    -
    \mathrm{sg}(P^{k})
    +
    P^{k},
\end{equation}
where \(\mathrm{sg}(\cdot)\) stops gradients. After symmetrization and self-loop removal, \(A_{\mathrm{dyn}}^{\mathrm{st}}\) re-encodes clean expression:
\begin{equation}
    Z_{\mathrm{dyn}}=f_{\mathrm{enc}}(X,A_{\mathrm{dyn}}^{\mathrm{st}}).
\end{equation}
The hard symmetrized graph updates \(A_{t+1}\) after each epoch. InfoNCE aligns current and dynamic views:
\begin{equation}
    \mathcal{I}(Z^a,Z^b)
    =
    -\frac{1}{n}\sum_{i=1}^{n}
    \log
    \frac{
    \exp(\mathrm{sim}(z_i^a,z_i^b)/\tau_c)
    }{
    \sum_{j=1}^{n}
    \exp(\mathrm{sim}(z_i^a,z_j^b)/\tau_c)
    },
\end{equation}
where \(\tau_c\) is the contrastive temperature. The contrastive losses are
\begin{equation}
    \mathcal{L}_{\mathrm{con}}^{pre}=\mathcal{I}(Z_m,Z_{\mathrm{dyn}}),
\end{equation}
\begin{equation}
    \mathcal{L}_{\mathrm{con}}^{fine}=\mathcal{I}(Z_c,Z_{\mathrm{dyn}}),
\end{equation}
with \(Z_c=f_{\mathrm{enc}}(X,A_t)\). DEC-style refinement supplies \(\mathcal{L}_{\mathrm{clu}}=\mathrm{KL}(P\|Q)\). The full objectives are
\begin{equation}
    \mathcal{L}_{\mathrm{pre}}=
    \lambda_r\mathcal{L}_{\mathrm{rec}}+
    \lambda_m\mathcal{L}_{\mathrm{mask}}+
    \lambda_z\mathcal{L}_{\mathrm{ZINB}}+
    \lambda_c\mathcal{L}_{\mathrm{con}}^{pre}.
\end{equation}
\begin{equation}
    \mathcal{L}_{\mathrm{fine}}=
    \lambda_{\mathrm{clu}}\mathcal{L}_{\mathrm{clu}}+
    \lambda_z\mathcal{L}_{\mathrm{ZINB}}+
    \lambda_c\mathcal{L}_{\mathrm{con}}^{fine}.
\end{equation}

\section{Experiments}
\subsection{Datasets and Evaluation Protocol}
We evaluate \modelname on 12 real scRNA-seq datasets from Adam\cite{Adam2017}, Klein\cite{Klein2015}, Plasschaert\cite{Plasschaert2018}, Tabula Muris\cite{Schaum2018}, Romanov\cite{Romanov2016}, Wang Lung\cite{Wang2018Dataset}, and Young\cite{Young2018}. They cover kidney, embryonic stem cells, trachea, limb muscle, diaphragm, heart, hypothalamus, and lung; Drop-seq, inDrop, 10x, and Smart-seq2; 2 to 11 cell types; 870 to 11269 cells; and 65.58\% to 94.70\% zero entries. We compare 10 baselines: graph-based scMGCA\cite{Yu2023scMGCA}, scAGC\cite{Li2025scAGC}, scCDCG\cite{Xu2024scCDCG}, scDSC\cite{Gan2022scDSC}, and scGNN\cite{Wang2021scGNN}, plus CIRCLE\cite{Wang2024CIRCLE}, scDML\cite{Yu2023scDML}, scziDesk\cite{Chen2020scziDesk}, scDeepCluster\cite{Tian2019scDeepCluster}, and scMAE\cite{Fang2024scMAE}. Metrics are NMI and ARI.

\subsection{Implementation Details}
All experiments are implemented in PyTorch and executed on an NVIDIA A6000 GPU with 48 GB memory. \modelname uses a hidden dimension of 256, a latent dimension of 128, an encoder dropout rate of 0.2, and TAKGConv with \(K=3\) and Fourier harmonic number \(G=4\). Pre-training runs for 1000 epochs with a learning rate of \(10^{-4}\). GDP-Mask uses a mask rate of 0.3, the dynamic graph neighbor number is \(k=15\), the Gumbel temperature is 1.0, the InfoNCE temperature is 0.7, and the loss weights are \(\lambda_r=1.0,\lambda_m=0.1,\lambda_z=1.0,\lambda_c=0.1\). Clustering refinement runs for 200 epochs with \(\lambda_{\mathrm{clu}}=1.0,\lambda_z=0.1,\lambda_c=0.01\), and the target distribution is updated every 8 epochs. Ablation and parameter sensitivity experiments follow the same training protocol and change only the module or parameter under analysis.

\subsection{Overall Clustering Performance}
Table~\ref{tab:comparison_experiments} reports mean scores over three seeds against ground-truth labels. Baselines marked with \(^{*}\) use unified reproduced results from scCluBench\cite{Xu2026scCluBench}; other baselines follow the same protocol. \modelname achieves the best average NMI/ARI of 0.8854/0.9105, ranking first on 9 of 12 datasets for NMI and 8 of 12 for ARI.

\begin{table*}[t]
    \centering
    \scriptsize
    \setlength{\tabcolsep}{1.8pt}
    \renewcommand{\arraystretch}{1.04}
    \caption{Overall clustering performance on 12 scRNA-seq datasets.}
    \label{tab:comparison_experiments}
    \begin{tabularx}{\textwidth}{@{}l>{\raggedright\arraybackslash}p{9.3em}|>{\centering\arraybackslash}X|>{\centering\arraybackslash}X>{\centering\arraybackslash}X>{\centering\arraybackslash}X>{\centering\arraybackslash}X>{\centering\arraybackslash}X|>{\centering\arraybackslash}X>{\centering\arraybackslash}X>{\centering\arraybackslash}X>{\centering\arraybackslash}X>{\centering\arraybackslash}X@{}}
    \toprule
    \textbf{Metric} & \textbf{Dataset} & \multicolumn{1}{c|}{\textbf{Ours}} & \multicolumn{5}{c|}{\textbf{GNN Methods}} & \multicolumn{5}{c}{\textbf{Deep Methods}} \\
    \cmidrule(lr){3-3}\cmidrule(lr){4-8}\cmidrule(lr){9-13}
     &  & \textbf{scKDGM} & \textbf{scMGCA} & \textbf{scAGC} & \textbf{scCDCG*} & \textbf{scDSC*} & \textbf{scGNN*} & \textbf{CIRCLE} & \textbf{scDML} & \textbf{scziDesk} & \makecell[c]{\textbf{scDeep}\\\textbf{Cluster*}} & \textbf{scMAE*} \\
    \midrule
    \multirow{13}{*}{NMI} & Adam & $\mathbf{0.9381}$ & $0.8664$ & $\underline{0.8871}$ & $0.6246$ & $0.7889$ & $0.7129$ & $0.6999$ & $0.8547$ & $0.8275$ & $0.6466$ & $0.7901$ \\
     & Klein & $\mathbf{0.8245}$ & $0.6939$ & $0.8004$ & $0.7753$ & $0.5748$ & $0.6492$ & $0.7652$ & $0.5957$ & $0.7855$ & $0.7323$ & $\underline{0.8134}$ \\
     & Plasschaert & $\underline{0.8438}$ & $0.7457$ & $\mathbf{0.8582}$ & $0.6559$ & $0.7690$ & $0.5419$ & $0.6400$ & $0.6793$ & $0.7952$ & $0.6047$ & $0.7198$ \\
     & Qx\_Limb\_Muscle & $\mathbf{0.9832}$ & $0.9417$ & $0.9402$ & $0.9412$ & $0.7970$ & $0.7760$ & $0.9050$ & $0.9470$ & $0.8987$ & $0.9112$ & $\underline{0.9682}$ \\
     & Qx\_Trachea & $\mathbf{0.8788}$ & $0.7018$ & $0.8335$ & $0.4984$ & $0.6750$ & $0.3636$ & $0.5736$ & $0.6716$ & $\underline{0.8473}$ & $0.5254$ & $0.8314$ \\
     & QS\_Diaphragm & $\mathbf{0.9765}$ & $0.9350$ & $0.9473$ & $0.8431$ & $0.9182$ & $0.9446$ & $0.8544$ & $0.8559$ & $0.9429$ & $0.9066$ & $\underline{0.9522}$ \\
     & QS\_Heart & $\mathbf{0.9199}$ & $0.9092$ & $\underline{0.9106}$ & $0.8465$ & $0.8681$ & $0.6814$ & $0.7671$ & $0.8465$ & $0.8334$ & $0.8070$ & $0.8644$ \\
     & QS\_Limb\_Muscle & $\underline{0.9637}$ & $0.9397$ & $\mathbf{0.9645}$ & $0.8779$ & $0.8847$ & $0.8736$ & $0.7709$ & $0.9470$ & $0.9587$ & $0.8946$ & $0.9511$ \\
     & QS\_Trachea & $\underline{0.7528}$ & $\mathbf{0.7710}$ & $0.6898$ & $0.6634$ & $0.5687$ & $0.7442$ & $0.5715$ & $0.6375$ & $0.6468$ & $0.6493$ & $0.6414$ \\
     & Romanov & $\mathbf{0.7800}$ & $0.6308$ & $0.6654$ & $0.6018$ & $0.5463$ & $0.5577$ & $0.6654$ & $0.5204$ & $\underline{0.7278}$ & $0.5755$ & $0.6931$ \\
     & Wang\_Lung & $\mathbf{0.9231}$ & $0.6516$ & $0.8786$ & $0.8347$ & $0.8526$ & $0.8252$ & $0.8571$ & $0.8177$ & $0.8192$ & $0.6748$ & $\underline{0.9184}$ \\
     & Young & $\mathbf{0.8406}$ & $0.8046$ & $0.7822$ & $0.6148$ & $0.7283$ & $0.4047$ & $0.7565$ & $\underline{0.8210}$ & $0.7537$ & $0.4931$ & $0.6790$ \\
     & AVG & $\mathbf{0.8854}$ & $0.7993$ & $\underline{0.8465}$ & $0.7315$ & $0.7476$ & $0.6729$ & $0.7355$ & $0.7662$ & $0.8197$ & $0.7018$ & $0.8185$ \\
    \midrule
    \multirow{13}{*}{ARI} & Adam & $\mathbf{0.9502}$ & $0.8725$ & $\underline{0.9028}$ & $0.4537$ & $0.7484$ & $0.5966$ & $0.6298$ & $0.8506$ & $0.8033$ & $0.4575$ & $0.7378$ \\
     & Klein & $\underline{0.8174}$ & $0.6886$ & $\mathbf{0.8188}$ & $0.7533$ & $0.5728$ & $0.5771$ & $0.7723$ & $0.6128$ & $0.8096$ & $0.7010$ & $0.7595$ \\
     & Plasschaert & $\underline{0.9017}$ & $0.7718$ & $\mathbf{0.9144}$ & $0.6399$ & $0.8320$ & $0.4125$ & $0.4837$ & $0.6563$ & $0.8495$ & $0.4255$ & $0.6162$ \\
     & Qx\_Limb\_Muscle & $\mathbf{0.9902}$ & $0.9572$ & $0.9525$ & $0.9660$ & $0.7820$ & $0.7541$ & $0.8695$ & $0.9448$ & $0.9059$ & $0.9332$ & $\underline{0.9836}$ \\
     & Qx\_Trachea & $\mathbf{0.9670}$ & $0.5404$ & $0.9340$ & $0.4659$ & $0.7278$ & $0.3002$ & $0.3621$ & $0.6683$ & $0.9201$ & $0.3730$ & $\underline{0.9456}$ \\
     & QS\_Diaphragm & $\mathbf{0.9872}$ & $0.9612$ & $\underline{0.9747}$ & $0.9184$ & $0.9416$ & $0.9699$ & $0.8550$ & $0.8272$ & $0.9668$ & $0.9353$ & $0.9714$ \\
     & QS\_Heart & $\mathbf{0.9615}$ & $0.9521$ & $\underline{0.9580}$ & $0.9162$ & $0.9260$ & $0.5544$ & $0.6131$ & $0.8303$ & $0.7544$ & $0.6928$ & $0.7757$ \\
     & QS\_Limb\_Muscle & $\underline{0.9808}$ & $0.9701$ & $\mathbf{0.9816}$ & $0.9315$ & $0.9305$ & $0.8833$ & $0.6302$ & $0.9669$ & $0.9723$ & $0.9097$ & $0.9718$ \\
     & QS\_Trachea & $\underline{0.8396}$ & $\mathbf{0.8686}$ & $0.8133$ & $0.5995$ & $0.5451$ & $0.7807$ & $0.4103$ & $0.5528$ & $0.7065$ & $0.4850$ & $0.5389$ \\
     & Romanov & $\mathbf{0.7969}$ & $0.5303$ & $0.6330$ & $0.5861$ & $0.4175$ & $0.5140$ & $0.6105$ & $0.4063$ & $\underline{0.7507}$ & $0.4795$ & $0.6705$ \\
     & Wang\_Lung & $\mathbf{0.9690}$ & $0.7288$ & $0.9394$ & $0.9134$ & $0.9257$ & $0.8975$ & $0.9320$ & $0.8988$ & $0.9023$ & $0.6637$ & $\underline{0.9667}$ \\
     & Young & $\mathbf{0.7644}$ & $0.7078$ & $0.6946$ & $0.4166$ & $0.6675$ & $0.1907$ & $0.7085$ & $\underline{0.7163}$ & $0.6448$ & $0.2618$ & $0.5162$ \\
     & AVG & $\mathbf{0.9105}$ & $0.7958$ & $\underline{0.8764}$ & $0.7134$ & $0.7514$ & $0.6193$ & $0.6564$ & $0.7443$ & $0.8322$ & $0.6098$ & $0.7878$ \\
    \bottomrule
    \end{tabularx}
    \vspace{1mm}
\end{table*}

\subsection{Ablation Study}
\begin{figure}[!ht]
    \centering
    \includegraphics[width=\linewidth]{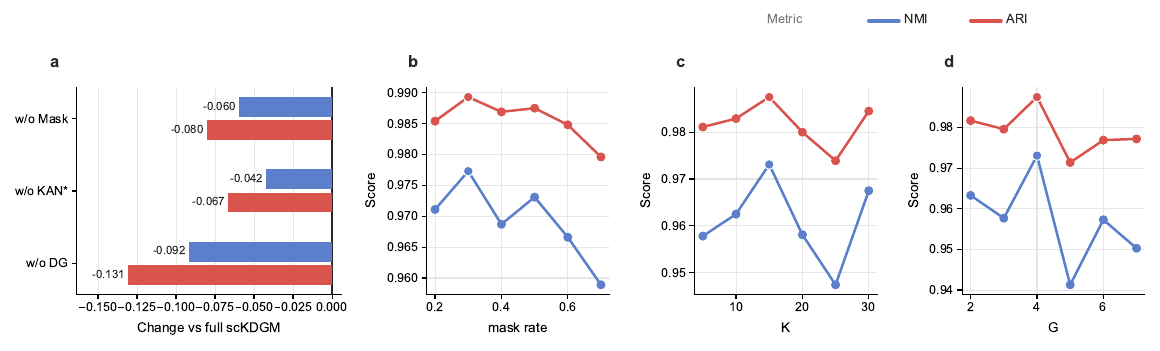}
    \caption{Ablation and parameter sensitivity results. Panel (a) reports average performance changes relative to full \modelname, where the TAGCN variant uses an encoder with a comparable parameter count. Panels (b)-(d) report sensitivity to mask rate, graph neighbor number \(k\), and Fourier harmonic number \(G\) on Quake Smart-seq2 Diaphragm.}
    \label{fig:ablation}
\end{figure}

We compare full \modelname with three variants. w/o Mask removes GDP-Mask pre-training and recovery supervision; w/o KAN replaces TAKGConv with a comparable TAGCN encoder of about 9M parameters; w/o DG disables dynamic graph updates. Fig.~\ref{fig:ablation} shows that w/o DG drops average NMI/ARI by 0.0919/0.1309, confirming the value of recovery-driven graph updates for reducing information bottlenecks, uneven propagation, and oversmoothing in static KNN graphs. w/o Mask drops NMI/ARI by 0.0596/0.0804, supporting graph-aware perturbation and observed-change recovery. w/o KAN also declines, showing that Fourier KAN transformations improve nonlinear high-order neighborhood aggregation.

\FloatBarrier

\subsection{Dynamic Graph Study}
\begin{figure}[!ht]
    \centering
    \includegraphics[width=\linewidth]{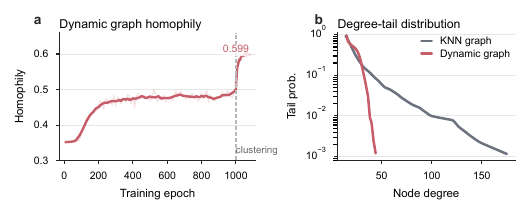}
    \caption{Graph-structure diagnostics on the Quake Smart-seq2 Diaphragm dataset. (a) Training evolution of dynamic graph edge homophily. (b) Degree-tail distributions of the KNN-constructed graph and the learned dynamic graph. The two graphs are constructed with the same neighbor number \(k\).}
    \label{fig:graph_quality}
\end{figure}

We further inspect cell type consistency and degree distribution. Edge homophily is the fraction of edges linking cells of the same type. In Fig.~\ref{fig:graph_quality}a, dynamic-graph homophily rises during training, reaches about 0.599 in refinement, and peaks at about 0.611, indicating increasingly cell-type-consistent edges. Fig.~\ref{fig:graph_quality}b shows that the KNN graph has a long-tailed hub structure, with maximum degree 175 and 99\% quantile 96.3. The dynamic graph reduces them to 44 and 37.0, suppressing hub concentration and mitigating uneven propagation, semantic mixing, and oversmoothing.

\FloatBarrier

\subsection{Parameter Sensitivity}
As shown in Fig.~\ref{fig:ablation}, the mask rate reaches the best NMI/ARI at 0.3 on Quake Smart-seq2 Diaphragm, with 0.9773/0.9893, and remains stable from 0.3 to 0.5. Thus moderate perturbation supplies useful self-supervision, while excessive masking weakens cell-specific patterns. For the dynamic graph, \(k=15\) gives 0.9731/0.9875; smaller \(k\) lacks coverage, and larger \(k\) increases semantic mixing. For Fourier harmonics, \(G=4\) is best with 0.9731/0.9875, balancing basis capacity and stability. The defaults \(k=15\), \(G=4\), and mask rate 0.3 therefore lie in the stable high-performance range.

\FloatBarrier
\section{Conclusion}
This paper presents \modelname, a KAN-guided dynamic graph masked learning framework for scRNA-seq clustering. GDP-Mask builds graph-aware distribution preserving perturbations, TAKGCN learns nonlinear graph-aware representations, and the mask-recovered feature matrix updates the dynamic graph, closing the loop between expression recovery and topology optimization. On 12 real datasets, \modelname achieves higher average NMI and ARI than representative deep and graph clustering methods. Ablation and sensitivity studies confirm the contributions of dynamic graph updates, GDP-Mask, and TAKGConv. However, pairwise dynamic graph construction needs approximate neighbor search or mini-batch updates for million-cell data, and biological interpretation should be further validated by marker enrichment and cell type annotation consistency. Future work will improve graph efficiency and extend \modelname to batch correction, multi-omics clustering, and spatial transcriptomics.

\section*{Acknowledgment}

This work was supported in part by the National Key R\&D Program of China under Grant 2025YFC3409000.

\bibliographystyle{IEEEtran}
\bibliography{references}

\end{document}